%% file: main.tex
\documentclass[10pt,twocolumn]{article}
\usepackage[a4paper,margin=0.8in,columnsep=0.25in]{geometry}
\usepackage{times}
\usepackage{graphicx,booktabs,amsmath,amssymb,xcolor,microtype}
\usepackage[numbers,sort&compress]{natbib}
\usepackage[pdftitle={Do Image Editors Follow Depth-Dependent Blur and Aperture Response? A Rendered-Ground-Truth Pilot Audit},colorlinks=true,linkcolor=blue!50!black,citecolor=blue!50!black,urlcolor=blue!50!black]{hyperref}
\usepackage[capitalise]{cleveref}
\usepackage{placeins}

\title{Do Image Editors Follow Depth-Dependent Blur and Aperture Response?\\A Rendered-Ground-Truth Pilot Audit}
\input{authors}
\date{}
\begin{document}
\twocolumn[\maketitle\begin{abstract}\input{sections/0_abstract}\end{abstract}\vspace{1em}]
\input{sections/1_introduction}
\input{sections/2_related_work}

\input{sections/3_protocol}
\input{sections/4_results}

\input{sections/5_conclusion}
\FloatBarrier
{\small\bibliographystyle{plainnat}\bibliography{references}}
\appendix
\input{sections/A_appendix}
\end{document}

%% file: authors.tex
\author{Zhihan Chen, Yuhuan Zhao, Yijie Zhu, Xinyu Yao,\\Mengcong Ren, Yuchen Sun, Yunqing Chen}

%% file: sections/0_abstract.tex
\noindent General image editors are asked to make a photo look as if it were taken at f/1.4, yet it is rarely checked whether the blur they add follows thin-lens optics.
A physical aperture edit spreads blur across depth in thin-lens proportions and changes the blur when the aperture changes; prior evaluations check blur monotonicity, sharpness-trend correlation, effective-aperture error, or vision-language judgments, and none we found reports the two properties separately at known depths.
In this pilot audit of two editors (Gemini~3.1 Flash Image and GPT-image-2.5) we compare against a rendered oracle: Blender Cycles scenes with true thin-lens depth of field, one blur-width estimator applied identically to oracle and editor outputs, and preregistered depth and aperture indices.
In 24 texture scenes rendered in one three-panel geometry, accepted and measurable panels show f/1.4-to-f/2.8 width ratios, $\bar\sigma_{1.4}/\bar\sigma_{2.8}$, of 0.99--1.18 against 1.98--2.13 for the oracle, and ratios of pooled median Gaussian-equivalent near/far blur widths of about 1.18--1.25 (Gemini) and 0.97--1.05 (GPT-image) against 1.69--1.82.
Preregistered black-box interventions show that qualitative wording changes blur strength by roughly 2--10 times, whereas a request for 2 versus 6 px changes it 1.1--1.3 times and no tested wording of the f-number meets the registered ``followed'' criterion.
The depth compression appears in the original and reversed centre-focus layouts; with the focus on the near panel, the available-panel depth index reaches the registered threshold, and the aperture response stays attenuated in every layout tested.
Scalar metrics adapted from published ones give oracle-like scores to synthetic editors whose proportions are compressed.
This is a pilot report on rendered scenes and two editors; the next stage tests real photographs, further editors, the validity of the indices, and a perceptual check.

%% file: sections/1_introduction.tex
\section{Introduction}
``Make it look like f/1.4'' is a routine request to a general-purpose image editor, and the editors return plausible bokeh.
Two properties define a physical aperture edit: how the blur is \emph{proportioned} across depth, and by how much it \emph{responds} when the aperture changes.
Existing evaluations check that blur grows as the requested aperture shrinks (Blur Monotonicity \citep{shen2025defocus}), correlate global sharpness trends (Bokeh Diffusion \citep{fortes2025bokeh}), report an effective-aperture error (CamEdit \citep{qin2025camedit}), or ask a vision-language model to judge realism \citep{picabench2025,phyeditbench2026}.
None of these jointly measures both at known depths; our audit reports depth proportion, aperture-response magnitude, and blur strength as three separate quantities against a measured, rendered thin-lens oracle.

\begin{figure*}[t]
  \centering
  \includegraphics[width=0.95\textwidth]{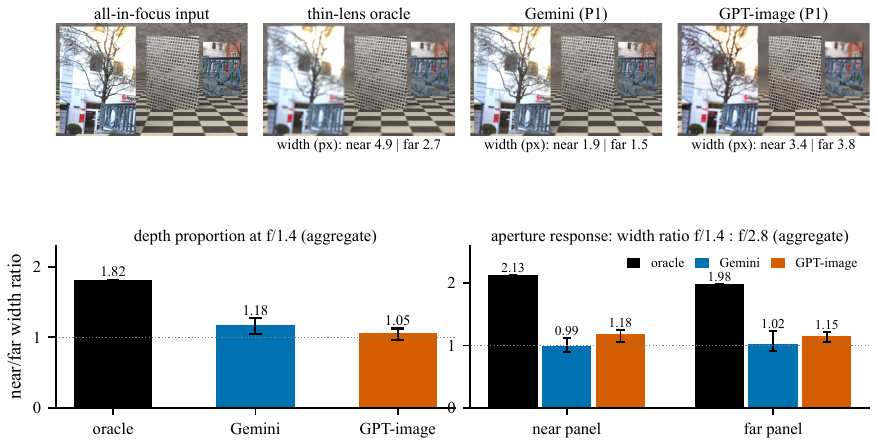}
  \caption{Top: one example scene (\texttt{r25\_X}, the first manifest job), requested f/1.4, prompt P1; each oracle and editor crop is labelled with the measured Gaussian-equivalent blur width $\sigma$ (px at 1536\,px image width) of the near (left, 1.5\,m) and far (right, 3.0\,m) panel; the input crop is not labelled. Bottom left: ratio of separately pooled near and far median widths at f/1.4; bottom right: ratio of pooled median widths at f/1.4 to f/2.8 for the near and far panel. Bottom panels aggregate 24 texture scenes $\times$ 2 content layouts (48 retained outputs per editor--prompt--aperture stratum) with prompt P1, using accepted and measurable panels; whiskers: 95\% bootstrap over scenes, conditional on the measured oracle and not capturing generation variability. Panel textures are crops of RealBokeh images \citep{seizinger2025bokehlicious} (CC BY-NC-SA 4.0).}
  \label{fig:hero}
\end{figure*}

In one three-panel geometry with 24 texture scenes and two editors, the audit finds that the f/1.4-to-f/2.8 width ratio, $\bar\sigma_{1.4}/\bar\sigma_{2.8}$, is 0.99--1.18 for the editors against 1.98--2.13 for the oracle, and that the ratio of pooled median Gaussian-equivalent near/far blur widths is about 1.18--1.25 (Gemini 3.1 Flash Image) and 0.97--1.05 (GPT-image-2.5) against 1.69--1.82 (\cref{fig:hero}, \cref{tab:main}).
The aperture attenuation holds when every missing panel is given its most favourable width; assigning those widths changes Gemini's categorical depth classification.
Ten preregistered interventions on six scenes each identify which tested prompt inputs change blur strength: qualitative words change it substantially, whereas the f-number in four wordings and a numeric blur size do not meet the registered criteria (\cref{tab:interventions}).
Depth compression appears in the original and reversed centre-focus layouts; near-focus indices computed from the available panels exceed the registered lower threshold (\cref{tab:v3}).
In short, the tested editors adjust blur strength in response to qualitative wording, little in response to an f-number or a pixel size, and in the centre-focus layouts do not distribute the blur across depth in thin-lens proportions.

\paragraph{Optics.}
Opening the aperture from f/2.8 to f/1.4 doubles the ideal defocus diameter at every off-focus depth, while the relative blur widths across depth are fixed by the scene geometry and the focus distance.
For a thin lens of focal length $f$ focused at distance $z_f$ (the lens model of \citealp{potmesil1981lens}), the circle-of-confusion diameter at scene distance $z$ and f-number $N$ is
\begin{equation}
c(z,N)=\frac{f^{2}}{N}\,\frac{z_f}{z_f-f}\,\Bigl|\frac{1}{z}-\frac{1}{z_f}\Bigr|,
\end{equation}
so it scales with $1/N$ for a fixed scene and with $|1/z-1/z_f|$ across depth.

\paragraph{Approach.}
We render scenes in Blender Cycles with true thin-lens depth of field: three equal-size textured panels at 1.5, 2.2, and 3.0\,m, the middle one in focus (\cref{sec:protocol}).
The all-in-focus render is the editor input, and renders at f/1.4, f/2.8, and f/5.6 are the oracle.
A local blur-width estimator measures each panel of each editor output and of each oracle render with the same procedure.
Two normalized indices, set in advance together with their cut-offs, report depth proportion and aperture response (1 means oracle behaviour, 0 means none).

\paragraph{Contributions.}
(1) An audit protocol that separates depth proportion, aperture response, and blur strength against a rendered thin-lens oracle.
(2) Control tests and a calibration grid of the measurement, including its biases for weak blur, disc kernels, and delivery at a smaller size.
(3) A preregistered pilot of two editors with attenuated aperture response and compressed depth proportions in this setting, with the sensitivity to missing panels reported.
(4) Ten preregistered black-box interventions identifying which prompt inputs change the blur, and a comparison with published-style scalar metrics on synthetic editors of known behaviour.

\paragraph{Scope and outline.}
This report is a pilot: the rendered-scene stage of the audit, with two editors.
It introduces the protocol and reports registered endpoints on rendered scenes; the next stage tests transfer to real photographs and broader editor coverage.
\Cref{sec:protocol} defines the scenes, the instrument, and the registered endpoints; \cref{sec:results} reports the measurement controls, the main audit, the layout and prompt interventions, and the comparison with scalar metrics; \cref{sec:limits} states what remains open and which follow-up test each open point calls for.

%% file: sections/2_related_work.tex
\section{Related Work}
\paragraph{Aperture and defocus control.}
Generative and editing methods give users a blur or aperture knob.
Bokeh Diffusion conditions a text-to-image model on a physical defocus parameter while keeping the scene fixed \citep{fortes2025bokeh}; the Adobe model of \citet{shen2025defocus} conditions on EXIF aperture and routes the output through a lens model; Generative Refocusing refocuses a single image \citep{mu2025refocus}; CamEdit edits aperture, focal plane, and shutter speed from text \citep{qin2025camedit}; AnyBokeh conditions on source and target circles of confusion \citep{anybokeh2026}; LensStyle adds lens-style rendering \citep{lensstyle2026}.
RealBokeh supplies paired real-image data for learning and evaluation \citep{seizinger2025bokehlicious}, Dr.Bokeh evaluates an occlusion-aware renderer on a ray-traced benchmark \citep{drbokeh2024}, and a bokeh-rendering stage supports metric depth estimation in \citet{zhang2025bokehdepth}.
Their evaluations measure different properties.
The Adobe work tests ordering: it counts the share of instances where signal energy falls as the aperture value falls \citep{shen2025defocus}.
Bokeh Diffusion tests global trend: it correlates Laplacian-variance trends with a reference \citep{fortes2025bokeh}.
CamEdit tests response magnitude in aggregate: it converts an edge-based blur estimate into an effective aperture through a simplified thin-lens model and reports the error against the target, including for general editing baselines \citep{qin2025camedit}.
Generative Refocusing compares one example against Nano Banana Pro qualitatively and notes that the editor also changes the content of the scene \citep{mu2025refocus}.
None of these evaluations separates the near/far proportion of blur from the overall response, and none reports blur strength relative to a rendered oracle at known depths.
Our endpoints report oracle-normalized depth proportion and aperture response separately, alongside blur strength.
Image-quality benchmarking of computational bokeh predates general instruction editors \citep{hauser2018bokeh}; the endpoints here are specific to depth proportion and aperture response against a rendered oracle.
Blind estimators of defocus from a single image \citep{zhuo2011defocus} and depth-from-defocus methods \citep{subbarao1994dfd} recover blur or depth from the image alone; our estimator is reference-based, because the all-in-focus render of each scene is known.

\paragraph{Audits of editors against physics and cameras.}
3DLP compares editors with real light-probe captures and finds the best ones remarkably consistent with real lighting \citep{kuechler2026lighting}.
\citet{rueckerl2026camera}, posted shortly before this report, read the implicit camera of editors (pitch, roll, focal length) from painted checkerboards on rendered cameras with exact ground truth; their rendered cameras have no depth of field.
Our audit applies the same logic of rendered ground truth and quantitative readout to defocus.
PICABench and PhyEditBench score physical realism with vision-language judges across categories such as light propagation, reflection, and refraction \citep{picabench2025,phyeditbench2026}, SpatialEdit measures spatial edits against Blender ground truth \citep{spatialedit2026}, and HP-Edit includes bokeh cases in a human-preference setting \citep{hpedit2026}.
This audit measures depth-dependent Gaussian-equivalent width ratios and oracle-normalized aperture response in general instruction editors.

%% file: sections/3_protocol.tex
\section{Audit Protocol}\label{sec:protocol}
The audit has three parts: rendered scenes with a thin-lens oracle, an instrument that reads the blur width of each panel, and registered endpoints with decision rules.
\paragraph{Scenes and oracle.}
Each scene shows three vertical panels of equal physical size ($0.351\times0.492$\,m) on a textured floor in front of a textured wall.
The panels stand at 1.5\,m (left, near), 2.2\,m (centre), and 3.0\,m (right, far) from a 50\,mm camera with a 36\,mm-wide sensor; images are $1536\times1024$.
The camera focuses on the centre panel.
We render with Blender Cycles (128 samples, no denoising).
The editor input is the all-in-focus render (depth of field off); the oracle renders use true thin-lens depth of field at f/1.4, f/2.8, and f/5.6.
Panel textures are two crops per scene from RealBokeh f/22 images \citep{seizinger2025bokehlicious} (CC BY-NC-SA 4.0, used for non-commercial research); the centre panel is a metal mesh.
A scene appears in two content layouts: crop $a$ near and crop $b$ far, or the reverse.
This counterbalances content across depth.
We use 24 scenes, hence 48 jobs.

\paragraph{Editors and prompts.}
We test \texttt{gemini-3.1-flash-image} (returned at $1264\times848$) and \texttt{gpt-image-2.5-sunburst} at high quality ($1536\times1024$).
Prompt P1 states the distances and asks for a 50\,mm lens at f/$N$ focused on the centre panel while keeping every panel's content; prompt P2 adds the literal cue ``blur must grow with distance from the focal plane'' with the 0.7\,m and 0.8\,m offsets (appendix); thin-lens blur is linear in $1/z$, not in distance, so P2 is a literal cue rather than a physically exact one.
Every job is edited at three requested apertures with both prompts, giving 288 cells per editor, plus 12 ``return unchanged'' images per editor that measure each editor's preservation floor.
We retain one output per cell.

\paragraph{Instrument.}
For each panel we warp the editor or oracle output to the all-in-focus reference with ECC affine alignment \citep{evangelidis2008ecc} and Lanczos resampling, and accept the alignment if the correlation is at least 0.90, both singular values of the linear part lie in $[0.94,1.06]$, shear is at most 0.03, and there is no reflection.
Inside an accepted panel, a non-blind estimator evaluates 41\,px windows centred at every interior pixel and finds the Gaussian width from a bank of blurred copies of the sharp reference that best matches the high-passed output, and calls a window measurable if its peak correlation is at least 0.80.
The panel width $\sigma$ is the median over measurable positions, defined when at least 30 are measurable, and reported in pixels at $1536$\,px image width.
Rejected panels, accepted panels without measurable windows, and measurable panels are counted separately.
The oracle widths are measured with the same estimator on the oracle renders and pooled by median.

\paragraph{Endpoints and decision rule.}
Let $\bar\sigma_N(p)$ denote the pooled median width of panel $p$ at requested aperture f/$N$, $R_N=\bar\sigma_N(\text{near})/\bar\sigma_N(\text{far})$, and let a circle denote the same quantity on the oracle.
The \emph{depth index} and the \emph{aperture index} are
\begin{align}
D&=\frac12\sum_{N\in\{1.4,2.8\}}\frac{\ln R_N}{\ln R^{\circ}_N},\\
A&=\frac12\sum_{p\in\{\text{near},\text{far}\}}\frac{\ln\bigl[\bar\sigma_{1.4}(p)/\bar\sigma_{2.8}(p)\bigr]}{\ln\bigl[\bar\sigma^{\circ}_{1.4}(p)/\bar\sigma^{\circ}_{2.8}(p)\bigr]}.
\end{align}
Both equal 1 for oracle behaviour and 0 when the editor shows no depth contrast (or no change with aperture) on average; they can be negative or exceed 1, and an average of zero does not imply zero response in each component.
Blur strength $r=\bar\sigma/\bar\sigma^{\circ}$ is reported alongside.
We exclude f/5.6 from the indices because the far-panel oracle width (about 0.4\,px) lies at the estimator's floor.
Intervals come from a bootstrap over the 24 scenes (4000 draws) and condition on the measured oracle.
Before any editor call on these stimuli we registered one-sided cut-offs on the interval of each index: ``replicated'' (the editor departs from optics) if the upper bound is at most 0.5, and ``compliant'' if the lower bound is at least 0.7; anything between is inconclusive.
``Compliant'' is a lower threshold, not an equivalence test against 1.
The registration, its execution notes, and its corrections are in the appendix.

%% file: sections/4_results.tex
\section{Results}\label{sec:results}
The results answer four questions in order: does the measurement behave on controls with known blur (\cref{fig:controls}); do the editors reproduce thin-lens proportion and aperture response (\cref{tab:main}); does the proportion compression depend on layout or focal plane (\cref{tab:v3}); and which prompt inputs change the blur, and do scalar metrics see it (\cref{tab:interventions,tab:metrics}).
\subsection{Measurement controls}
\begin{figure*}[t]
  \centering
  \includegraphics[width=0.95\textwidth]{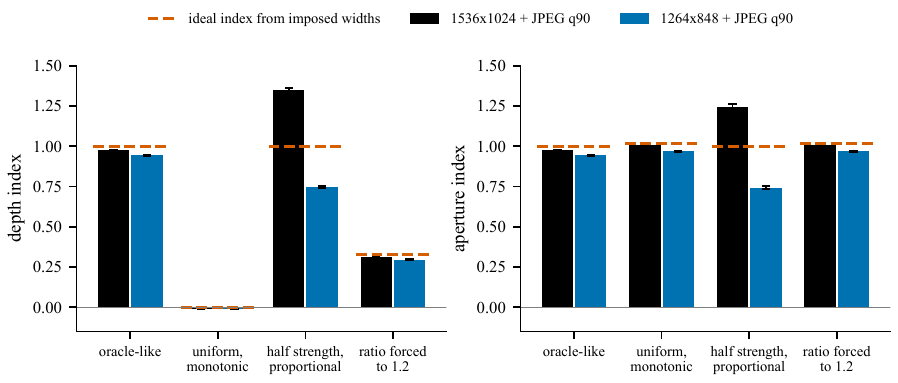}
  \caption{Exploratory control test of the instrument: synthetic "editors" with known Gaussian blur on the panels, delivered at native size or at Gemini's output size with JPEG, scored with the same pipeline. Dashed: ideal index computed from the imposed widths and the measured oracle (0.327 for the forced 1.2 ratio; aperture index 1.017 for the two mean-width controls). Both delivery conditions include JPEG q90. Indices use f/1.4 and f/2.8: 24 scenes $\times$ 2 layouts = 96 synthetic images per control and delivery condition. The half-strength control departs from ideal recovery in both delivery conditions (native 1.35/1.25, resized 0.75/0.74). Not shown: sharp-only and wrong-focus controls (native indices undefined; resized sharp-only $-0.054$/0, wrong-focus $-4.58$/0.35).}
  \label{fig:controls}
\end{figure*}

We assessed the measurement with exploratory synthetic ``editors'' that blur the panels of the all-in-focus renders with known widths taken from the oracle, delivered at native size or at Gemini's output size, both with JPEG compression (\cref{fig:controls}).
An oracle-like control returns depth and aperture indices of 0.98 and 0.97 (native) and 0.94 and 0.94 (Gemini size).
Uniform blur that shrinks with the aperture returns a depth index near 0 and an aperture index near 1, so the pair separates ``uniform but monotonic'' from optics, a failure that ordering metrics accept.
Forcing the near/far ratio to 1.2 returns a depth index of 0.31 (native) and 0.29 (Gemini size), against an ideal of 0.33.
A half-strength control with correct proportions departs from the ideal of 1 in both delivery conditions (native 1.35 and 1.25; Gemini size 0.75 and 0.74).
Weak-blur recovery depends on delivery size.
The absolute scale of the Gaussian-equivalent width relative to the physical blur is uncertain by about 25\% (an analytic and a slanted-edge check, appendix); the indices are ratios of readings of the same estimator, so they cancel a common scale factor but not a bias that depends on blur strength or delivery size (the half-strength control below).

\subsection{Depth proportion}
\begin{table*}[t]
\centering\footnotesize\setlength{\tabcolsep}{3.5pt}
\caption{Pilot audit in one synthetic three-panel geometry: 24 texture scenes $\times$ 2 content layouts (48 outputs) per editor--prompt--aperture stratum, one retained output per scene--layout--prompt--aperture cell. Widths are median Gaussian-equivalent blur widths (px at 1536\,px) of accepted, measurable panels; ratio = near/far; depth and aperture indices (1 = oracle, 0 = no response) pool f/1.4 and f/2.8 (aperture index: f/1.4 vs f/2.8). 95\% intervals come from a bootstrap over scenes, conditional on the measured oracle; they do not capture generation variability. Last column: panels accepted and measurable at f/1.4 and f/2.8 (out of 96 = 48 jobs $\times$ 2 apertures); the rest are rejected by the alignment rule or unmeasurable. Ratios are computed from the pooled medians before display rounding. Gemini returns 1264$\times$848 JPEG, the oracle and GPT-image 1536$\times$1024 PNG.}
\label{tab:main}
\begin{tabular}{llrrrrrrccc}
\toprule
 & & \multicolumn{3}{c}{f/1.4} & \multicolumn{3}{c}{f/2.8} & depth index & aperture index & (of 96) \\
\cmidrule(lr){3-5}\cmidrule(lr){6-8}
Source & Prompt & near & far & ratio & near & far & ratio & & & usable panels \\
\midrule
oracle & -- & 5.0 & 2.8 & 1.82 & 2.4 & 1.4 & 1.69 & 1 & 1 & 90/96 far, 96/96 near \\
Gemini & P1 & 1.9 & 1.7 & 1.18 & 2.0 & 1.6 & 1.22 & 0.326 [0.174, 0.445] & 0.007 [-0.129, 0.204] & 85/96 far, 95/96 near \\
Gemini & P2 & 2.0 & 1.6 & 1.25 & 1.8 & 1.4 & 1.24 & 0.392 [0.269, 0.442] & 0.143 [0.069, 0.247] & 82/96 far, 96/96 near \\
GPT-image & P1 & 4.0 & 3.8 & 1.05 & 3.4 & 3.3 & 1.03 & 0.075 [-0.060, 0.175] & 0.214 [0.098, 0.264] & 90/96 far, 96/96 near \\
GPT-image & P2 & 4.5 & 4.6 & 0.97 & 4.1 & 4.1 & 1.00 & -0.022 [-0.153, 0.064] & 0.158 [0.033, 0.209] & 84/96 far, 96/96 near \\
\bottomrule
\end{tabular}
\end{table*}

\begin{figure}[t]
  \centering
  \includegraphics[width=0.95\columnwidth]{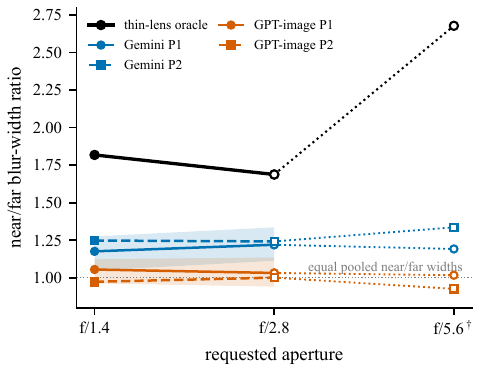}
  \caption{Near/far blur-width ratio versus requested aperture (solid: P1, dashed: P2; bands: 95\% scene bootstrap for P1). $^\dagger$f/5.6 is limited by the estimator floor (far-panel oracle width about 0.4\,px) and is descriptive only. 24 texture scenes (bootstrap clusters) $\times$ 2 content layouts = 48 retained outputs per editor--prompt--aperture stratum; endpoints use accepted, measurable panels; P1 states the distances, P2 adds a distance-from-focus cue.}
  \label{fig:ratio}
\end{figure}

\Cref{tab:main} and \cref{fig:ratio} give the main numbers.
At f/1.4 and f/2.8, the oracle's near/far width ratio is 1.82 and 1.69.
Gemini's pooled ratio is 1.18 (P1) and 1.25 (P2) at f/1.4 and 1.22 and 1.24 at f/2.8; GPT-image's is 1.05 and 0.97 at f/1.4 and 1.03 and 1.00 at f/2.8.
The editor-to-oracle ratios fall between 0.54 and 0.74.
The depth indices are 0.33 [0.17, 0.45] and 0.39 [0.27, 0.44] for Gemini and 0.08 [$-0.06$, 0.18] and $-0.02$ [$-0.15$, 0.06] for GPT-image (P1 and P2).
All four meet the registered ``replicated'' cut-off (upper bound at most 0.5) for these accepted and measurable panels.
Blur strength differs between editors: relative to the oracle, Gemini's near and far widths at f/1.4 under P1 are 0.39 and 0.60 times the oracle's, and GPT-image's are 0.79 and 1.36 times; GPT-image applies about the same width to both panels.
A control that forces a ratio of 1.2 returns an index of 0.31, close to Gemini's 0.33, so a pooled ratio near 1.2 corresponds to an index near 0.3.

\subsection{Aperture response}
\begin{figure}[t]
  \centering
  \includegraphics[width=0.95\columnwidth]{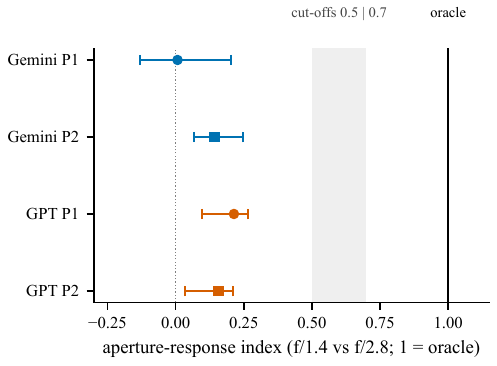}
  \caption{Aperture-response index for the requested f/1.4 versus f/2.8 change (1 = oracle response, 0 = none), averaged over the near and far panel log-response indices (each normalized by its own oracle response), with 95\% scene-bootstrap intervals. Registered rule: ``replicated'' if the upper bound $\le 0.5$, ``compliant'' if the lower bound $\ge 0.7$ (shaded band: 0.5--0.7). 24 texture scenes (bootstrap clusters) $\times$ 2 content layouts = 48 retained outputs per editor--prompt--aperture stratum; endpoints use accepted, measurable panels; P1 states the distances, P2 adds a distance-from-focus cue.}
  \label{fig:aperture}
\end{figure}

For the requested change from f/1.4 to f/2.8, the oracle's width ratio is 1.98--2.13 across the two side panels, and the editors' is 0.99--1.18 (\cref{fig:aperture}).
The aperture indices are 0.01 [$-0.13$, 0.20] and 0.14 [0.07, 0.25] for Gemini and 0.21 [0.10, 0.26] and 0.16 [0.03, 0.21] for GPT-image (P1 and P2).
The attenuated response also holds when every missing panel is assigned the most favourable width in the estimator's range, with the largest resulting upper bound at 0.33.

\subsection{Sensitivity to missing panels}
The depth index uses accepted panels with measurable windows; at the primary apertures 85 of 96 far panels are usable for Gemini P1, 82 for Gemini P2, 90 for GPT-image P1, and 84 for GPT-image P2, against 90 for the oracle (\cref{tab:main}).
When missing widths are assigned the most favourable value in the estimator's range, Gemini's depth upper bounds rise to 0.53 and 0.54, just above the registered cut-off, while GPT-image's depth result and both editors' aperture results stay below it.
Gemini's categorical depth classification is therefore sensitive to missing panels; its worst-case indices remain well below the 0.7 compliance cut-off.
As a focal-plane preservation check, the centre-panel threshold is $\max(1.0\,\mathrm{px},\text{P0 floor}+0.3\,\mathrm{px})$, which gives 1.24\,px for Gemini and 1.28\,px for GPT-image. Across all three apertures, Gemini P1 and P2 have 1 of 143 and 0 of 140 measurable centres above the threshold (1 and 4 centres unknown), and GPT-image P1 and P2 have 7 of 120 and 7 of 119 (24 and 25 unknown).

\subsection{Layout and focal plane}
We repeated the audit with a reversed layout (near panel on the right, focus on the centre panel), with the focus moved to the near panel, and with the original layout re-run under a prompt that states the sensor size and calls the input all-in-focus (\cref{tab:v3}); the cut-offs and a rule for generalization were fixed in an amended registration before any call.
In the reversed layout, both editors keep the compression: depth indices of 0.21 [0.07, 0.45] for Gemini and 0.09 [$-0.27$, 0.37] for GPT-image, and aperture indices of 0.21 [$-0.03$, 0.36] and 0.03 [$-0.07$, 0.21]; all four meet the ``replicated'' cut-off.
With the focus on the near panel, the blurred panels are the centre and far panels, and the oracle's centre-over-far ratio is 0.63 (f/1.4) and 0.64 (f/2.8).
The editors' ratios are 0.57 and 0.59 (Gemini) and 0.60 and 0.65 (GPT-image), and the depth indices are 1.18 [0.97, 1.41] and 1.02 [0.82, 1.22], which meet the ``compliant'' cut-off, while the aperture indices stay low at 0.17 [0.03, 0.30] and 0.04 [$-0.08$, 0.19].
By the preregistered rule, the depth compression therefore does not generalize to this layout: it appears when the near panel lies in front of the focal plane (original and reversed layouts) and not when the focus is on the near panel, whereas the aperture attenuation appears in all three.
These indices describe accepted and measurable panels, not complete edits, and in layout F the mesh centre panel is one of the two panels measured as blurred; the joint counts of usable panels are in the appendix.
In this layout, GPT-image's focal panel exceeds the 1\,px threshold in 16 of 48 outputs (Gemini: 4 of 48), and 32 of its 48 centre panels are rejected or fall below 20\% coverage, so its depth index rests on 16 centre panels (Gemini: 42).
The original layout under the corrected prompt gives ratios of 1.07 and 1.19 (Gemini) and 1.02 and 1.19 (GPT-image) against 1.80 and 1.70, with depth indices of 0.22 [0.02, 0.52] (Gemini; ``inconclusive'' because the upper bound exceeds 0.5) and 0.18 [$-0.11$, 0.32] (GPT-image; ``replicated'') on six scenes.

\begin{table*}[t]
\centering\footnotesize\setlength{\tabcolsep}{3.5pt}
\caption{Layout and focal-plane generalization (preregistered before any call; one retained output per scene--layout--aperture cell; 12 scenes $\times$ 2 content layouts for R and F, 6 scenes for the original layout re-run). Layout R: reversed positions, focus on the centre panel, depth index from near (right) over far (left). Layout F: focus on the near panel, depth index from the centre panel over the far panel. Original: layout of \cref{tab:main} with prompt P1s (states the sensor size, calls the input all-in-focus). Ratios at f/1.4 / f/2.8 are ratios of pooled median widths; indices use f/1.4 and f/2.8 with 95\% bootstrap intervals over scenes; panels with coverage below 20\% are excluded. Focus violations: focal-panel violations among accepted panels with a measurable width, without the coverage exclusion (unknown in brackets). Joint jobs: jobs of 24 with both blurred panels usable at both apertures (not tabulated for the original-layout re-run).}
\label{tab:v3}
\begin{tabular}{llcccccc}
\toprule
Editor & Layout & oracle ratio & editor ratio & depth index & aperture index & focus violations & joint jobs \\
\midrule
Gemini & R & 1.82/1.69 & 1.11/1.14 & 0.21 [0.07, 0.45] & 0.21 [-0.03, 0.36] & 2/30 (18) & 18 \\
Gemini & F & 0.63/0.64 & 0.57/0.59 & 1.18 [0.97, 1.41] & 0.17 [0.03, 0.30] & 4/48 (0) & 16 \\
Gemini & original & 1.80/1.70 & 1.07/1.19 & 0.22 [0.02, 0.52] & 0.21 [-0.16, 0.36] & 0/24 (0) & -- \\
GPT-image & R & 1.82/1.69 & 1.05/1.06 & 0.09 [-0.27, 0.37] & 0.03 [-0.07, 0.21] & 7/35 (13) & 20 \\
GPT-image & F & 0.63/0.64 & 0.60/0.65 & 1.02 [0.82, 1.22] & 0.04 [-0.08, 0.19] & 16/48 (0) & 1 \\
GPT-image & original & 1.80/1.70 & 1.02/1.19 & 0.18 [-0.11, 0.32] & 0.07 [-0.27, 0.22] & 3/17 (7) & -- \\
\bottomrule
\end{tabular}
\end{table*}
\subsection{What changes the blur}\label{sec:interventions}
To identify which inputs change the blur we changed one input at a time, except where noted, and measured the same panels (\cref{tab:interventions}; each block was registered before any call and uses six scenes; the minimal aperture wording also drops the stated distances and the focus instruction).
In short, qualitative words move the blur strength, a numeric blur size moves it little, and no aperture wording meets the registered ``followed'' criterion, with some outcomes inconclusive because their intervals are wide.

\begin{table*}[t]
\centering\scriptsize\setlength{\tabcolsep}{2.5pt}
\caption{Preregistered interventions (six scenes each; original layout and f/2.8 unless stated; the input image is unchanged unless stated). $\ln R$: change of the log left/right width ratio relative to the neutral prompt; ``$\times$'': ratio of pooled median widths (left / right panel); indices as in \cref{tab:main}; brackets: 95\% scene bootstrap. Labels are those the registered rule gives; $^{*}$the registered label is ``disproportional'' (ratio-change interval not within $\pm$0.15: Gemini [$-$0.49, 0.78], GPT-image [$-$0.09, 0.29]); both intervals include zero, so proportionality is unresolved. v11 widths use max(median, 0.4 px) in numerator and denominator (post-registration change for zero widths).}
\label{tab:interventions}
\begin{tabular}{@{}l@{\hspace{3pt}}p{4.6cm}lccp{3.3cm}@{}}
\toprule
Block & Intervention & Measure & Gemini & GPT-image & Registered label \\
\midrule
v5 & stated distances swapped & $\Delta\ln R$ & +0.11 [$-$0.15, 0.39] & $-$0.12 [$-$0.21, 0.10] & intermediate \\
v5 & stated distances 0.5 m / 20 m & $\Delta\ln R$ & +0.35 [0.02, 0.56] & $-$0.13 [$-$0.50, 0.29] & intermediate \\
v6 & ``slight'' $\to$ ``strong'' blur & width $\times$ & 2.2 / 1.9 & 3.0 / 2.9 & proportionality unresolved$^{*}$ \\
v9c & numeric blur size 2 $\to$ 6 px & width $\times$ & 1.13 / 1.10 & 1.30 / 1.14 & intermediate \\
v10 & f-number worded 3 ways (WS / WZ / WM) & aperture index & 0.08 / 0.14 / 0.48 & 0.14 / 0.30 / $-$0.06 & no wording meets ``followed'' \\
v9a & focus on far panel, f/1.4 vs f/2.8 & aperture index & 0.05 [$-$0.24, 0.33] & 0.15 [0.07, 0.31] & attenuated \\
v11 & ``very shallow'' vs ``very deep'' DoF & width $\times$ & 5.0 / 3.0 & 9.1 / 10.4 & words control the blur \\
v9b & floor and wall removed & $\Delta\ln R$ & $-$0.08 [$-$0.51, 0.04] & $-$0.11 [$-$0.25, 0.21] & inconclusive \\
v7 & both panels behind / in front of the focus & depth index & 0.48 / 0.34 & 0.47 / 0.29 & rejected (in front: compressed) \\
v8 & oracle ratio fixed at 2.6; opposite / both in front / both behind & depth index & 0.47 / 0.51 / 0.60 & $-$0.11 / 0.06 / 0.14 & Gemini inconclusive, GPT-image rejected \\
v4 & extreme distances G1 / G2 & depth index & 0.26 / 0.31 & 0.26 / 0.03 & inconclusive (both editors) \\
\bottomrule
\end{tabular}
\end{table*}
\paragraph{Numeric requests change the blur little.}
Four wordings of the f-number (the P1s baseline, a photography-terms wording, a Chinese wording, and a minimal ``Change the aperture to f/$N$'') give aperture indices of $-$0.06--0.30 for GPT-image and 0.08--0.48 for Gemini (the highest, Gemini with the minimal wording, is 0.48 [$-$0.31, 1.72], an interval wide enough to include the oracle response, so that outcome is inconclusive); no wording meets the registered ``followed'' criterion.
Moving the focus to the far panel leaves the response attenuated (0.05 and 0.15), and asking for a Gaussian blur of about 6 px instead of 2 px changes the median widths by 1.1--1.3 times, not by three.
Swapping the stated distances in the text (image unchanged) changes the log left/right width ratio by $+$0.11 [$-$0.15, 0.39] for Gemini and $-$0.12 [$-$0.21, 0.10] for GPT-image, which the registered rule labels intermediate for both; stating extreme distances (0.5 and 20 m) raises both side-panel widths for GPT-image (left 3.96 to 4.90 px, right 3.33 to 4.67 px) and the left/right ratio for Gemini.
\paragraph{Qualitative words have a large effect.}
Asking for ``a very slight'' versus ``a very strong'' amount of blur changes the side-panel widths by 1.9--3.0 times, and ``a very shallow'' versus ``a very deep depth of field'' by roughly 3 to 10 times (the oracle's f/1.4 over f/5.6 is 4.3 and 6.4 times; the v11 widths are floored at 0.4 px before taking ratios); under the deep wording both editors return almost sharp panels (0.0--0.5 px).
In these tests the amount of blur follows the words of the request more than its numerals.
\paragraph{Measurements contradict depth-rank and foreground-flattening predictions.}
With extreme distances rendered in the scene (block v4, far panel at 6\,m), both editors blurred the far panel more than the near one (ratio 0.55) when the far panel was much farther, which a depth-rank rule (equal blur for panels one plane from the focus) does not predict.
With the focus on the far panel the centre/near ratio was 0.48 (Gemini) and 0.63 (GPT-image), not near 1 as a foreground-flattening rule predicts.
In the matched-oracle test, with the oracle ratio fixed at 2.6, the three configurations give indices of 0.47, 0.51, 0.60 (Gemini) and $-$0.11, 0.06, 0.14 (GPT-image); the registered rule labels the straddle hypothesis inconclusive for Gemini and rejected for GPT-image, and with both panels on the same side the compression persists (\cref{tab:interventions}).
Removing floor and wall leaves the near/far ratio compressed (1.10 and 1.07 against 1.19, oracle about 1.75).
A post hoc summary of the compression as a power of the oracle ratio is in the appendix.

\subsection{Do scalar metrics see this?}\label{sec:metrics}
We implemented adaptations of three published-style scalar measures and applied them to the synthetic editors of \cref{sec:results} and to the editor outputs (\cref{tab:metrics}).
Monotonicity and trend correlation flag the weak aperture response of the editors (59--67\% of aperture pairs ordered over the whole frame and 57--59\% over the side panels, correlation 0.22--0.28, against 100\% and 0.99 for the oracle-like control), but they give oracle-like scores to the uniform-blur control and to the control with the ratio forced to 1.2.
A global strength error is small for those two controls and for Gemini at f/2.8 (0.09), although Gemini's depth proportion is compressed (depth index 0.33); it flags the half-strength control (0.78) and GPT-image (0.59).
Among the tested measures, the depth index distinguishes these proportion failures.

\begin{table}[t]
\centering\scriptsize\setlength{\tabcolsep}{2pt}
\caption{Published-style scalar metrics (our adaptations, not the cited implementations) next to the depth index $D$ and aperture index $A$. M1: share of consecutive apertures (f/1.4, 2.8, 5.6) with image energy increasing with the f-number (whole frame; side panels), \%; M2: correlation of the Laplacian-variance trend with the oracle's; M3: median absolute log error of a single global blur level at f/2.8. M1--M3 for the synthetic editors use 12 jobs at native size without JPEG; their $D$ and $A$ are those of the 48-job controls with JPEG q90 (\cref{fig:controls}); editors: 48 jobs, prompt P1.}
\label{tab:metrics}
\begin{tabular}{lcccccc}
\toprule
Source & M1w & M1p & M2 & M3 & $D$ & $A$ \\
\midrule
synth.: oracle-like & 100 & 100 & 0.99 & 0.02 & 0.98 & 0.97 \\
synth.: uniform & 100 & 100 & 0.99 & 0.01 & -0.01 & 1.01 \\
synth.: half strength & 100 & 100 & 0.99 & 0.78 & 1.35 & 1.25 \\
synth.: ratio 1.2 & 100 & 100 & 0.99 & 0.01 & 0.31 & 1.01 \\
Gemini (P1) & 59 & 59 & 0.28 & 0.09 & 0.33 & 0.01 \\
GPT-image (P1) & 67 & 57 & 0.22 & 0.59 & 0.08 & 0.21 \\
\bottomrule
\end{tabular}
\end{table}

%% file: sections/5_conclusion.tex
\section{Limitations and Conclusion}\label{sec:limits}
\paragraph{Limitations.}
(1) Three fixed panel distances and a flat-panel geometry; the layout, focal-plane and intervention tests use six to twelve scenes each, one aperture or two, and an equal-projected-size layout was used only in some blocks.
(2) The estimator reports an equivalent Gaussian width while real bokeh has a disc-shaped kernel; the calibration grid shows an accurate recovery of Gaussian blur above 1.5 px and an overestimate of 11--25\% for disc kernels, weak blur below about 1 px is not recovered reliably, the absolute scale of the Gaussian-equivalent width relative to the physical blur is uncertain by about 25\%, and the recovered indices of a proportional half-strength control depend on the delivery size (1.35 and 1.25 at native size, 0.75 and 0.74 at Gemini's size).
(3) Two editors, one retained output per cell, and bootstrap intervals over scenes that do not capture generation variability; f/5.6 sits at the estimator floor and is descriptive only.
(4) Panels are photographs mounted on boards in a rendered room, and whether the same behaviour appears in real captures with recorded distances is untested.
(5) The hypotheses of \cref{sec:interventions} were motivated by exploratory observations and their tests were preregistered; the compression-exponent summary in the appendix is post hoc.
(6) The scalar metrics of \cref{sec:metrics} are our adaptations of the published measures, and their columns in \cref{tab:metrics} use 12 synthetic jobs without JPEG while $D$ and $A$ use the 48-job controls with JPEG.

\paragraph{Practical reading.}
On the tested scenes, a request for stronger or weaker blur is better made with qualitative words (``very shallow depth of field, strong background blur'') than with an f-number or a pixel size, and in the centre-focus layouts tested the distribution of blur across depth should not be expected to follow thin-lens proportions.

\paragraph{Conclusion.}
In a three-panel geometry with 24 texture scenes and two editors, accepted and measurable panels show f/1.4-to-f/2.8 width ratios of 0.99--1.18 against 1.98--2.13 for the measured oracle, and near/far width ratios of 0.97--1.25 against 1.69--1.82.
On six scenes per block and two editors, preregistered interventions show that qualitative words about depth of field and blur change blur strength substantially, whereas the f-number (four wordings) and a numeric blur size do not meet the registered criteria; the tested geometric hypotheses did not explain the compression of proportions.
Separating proportion, response, and strength makes these failures visible as distinct quantities, which published-style scalar metrics partly miss, and gives general editors a rendered-ground-truth test for aperture edits.

\paragraph{Next stage.}
This pilot introduces an oracle-referenced protocol and reports rendered-scene findings for two editors.
The next stage tests transfer to tripod photographs with recorded distances and aperture brackets, broadens editor coverage with repeated generations, and quantifies index bias and missing-panel sensitivity across blur strengths, kernels, delivery sizes, and focal-plane layouts.
It will compare faithful implementations of published scalar metrics, release a public evaluation suite, and test whether blinded observers detect the measured differences.

%% file: sections/A_appendix.tex
\section{Appendix}
\Cref{tab:controls_all} lists all six measurement controls.
\begin{table*}[!t]
\centering\footnotesize\setlength{\tabcolsep}{4pt}
\caption{All six measurement controls (not preregistered). Depth and aperture indices with 95\% scene-bootstrap intervals, at native size ($1536\times1024$ + JPEG q90) and at Gemini's size ($1264\times848$ + JPEG q90); last column: median centre-panel width at native size (px). Indices are undefined when a required width ratio or logarithm is undefined, including zero widths.}
\label{tab:controls_all}
\begin{tabular}{lccccc}
\toprule
 & \multicolumn{2}{c}{native size} & \multicolumn{2}{c}{Gemini size} & centre \\
\cmidrule(lr){2-3}\cmidrule(lr){4-5}
Control & depth index & aperture index & depth index & aperture index & width \\
\midrule
oracle-like & 0.976 [0.973, 0.978] & 0.974 [0.968, 0.975] & 0.944 [0.942, 0.945] & 0.943 [0.940, 0.945] & 0.00 \\
uniform, monotonic & -0.009 [-0.011, -0.005] & 1.006 [1.002, 1.009] & -0.008 [-0.010, -0.006] & 0.968 [0.965, 0.970] & 0.00 \\
half strength, proportional & 1.348 [1.338, 1.360] & 1.245 [1.233, 1.260] & 0.747 [0.739, 0.754] & 0.743 [0.732, 0.752] & 0.00 \\
near/far ratio forced to 1.2 & 0.314 [0.312, 0.318] & 1.008 [1.005, 1.011] & 0.294 [0.291, 0.299] & 0.969 [0.965, 0.972] & 0.00 \\
focus on the near panel & undefined & undefined & -4.582 [-4.593, -4.570] & 0.349 [0.348, 0.350] & 2.46 \\
no blur & undefined & undefined & -0.054 [-0.073, -0.038] & 0.000 [0.000, 0.000] & 0.00 \\
\bottomrule
\end{tabular}
\end{table*}
\subsection{Prompts}
\textbf{P1:} ``This photo shows three flat panels standing in a row, about 1.5 m (left), 2.2 m (centre) and 3 m (right) from the camera. Edit it so it looks like it was taken with a 50mm lens at f/$N$ instead of f/22, from the same camera position, focused on the centre panel. Keep the composition and the content of every panel exactly as it is.''
\textbf{P2:} P1 followed by ``Blur must grow with distance from the focal plane: the left panel (0.7 m in front of it) and the right panel (0.8 m behind it) must both be blurred, the centre panel perfectly sharp.''
\textbf{P0:} ``Return this image unchanged.''
The input is an all-in-focus render that P1 and P2 describe as ``f/22'', and they do not state the sensor size; the follow-up in \cref{tab:v3} uses a prompt that gives the sensor size and describes the input as all-in-focus. The P2 cue is a linear distance cue, whereas the thin-lens relation is linear in $1/z$.

\subsection{Registration and execution record}
The pilot was registered before any editor call on these stimuli (scene generation, prompts, estimator, endpoints, decision rule, and spending limits; frozen text timestamped 2026-10-03T17:58:24Z, SHA-256 \texttt{3ae696d93c061eaa\allowbreak 9f78626d92649ecf\allowbreak b326ce2156d0abd5\allowbreak b46ff4c2f4b914d9}, before the execution notes were appended).
Later notes record that: the preservation controls use the first 12 jobs of the manifest; the main runs started on the jobs whose render had finished and continued on the rest with the same prompts and task order; and the robustness analyses (coverage, missing-panel sensitivity, control renders) are exploratory.
Gemini's log contains 426 attempts: 378 successes, of which 78 cells were generated twice because two runners overlapped (the later image is retained and the earlier one was overwritten), and 48 HTTP 402 failures after the prepaid credits ran out, which produced no image.
This leaves 300 retained images per editor (288 main cells and 12 preservation controls).
GPT-image's log contains 300 attempts, all successful.
We report one retained output per scene--layout--prompt--aperture cell.
Adapter-estimated costs were US\$25.33 for Gemini and US\$16.22 for GPT-image, US\$41.55 in total, above the registered US\$38 cap (the Gemini cap had been amended from US\$20.00 to US\$20.20; the total cap was unchanged); these are logged estimates, not invoices.

\subsection{Layout and focal-plane run}
The registration of this follow-up (12 scenes, two layouts, f/1.4 and f/2.8, prompt P1s) was frozen before any call and amended twice before the first call: the decision rule was tightened (a layout counts toward generalization only if both layouts meet the ``replicated'' cut-off), a re-run of 12 original-layout jobs was added, and the spending caps were set to US\$8.30 (Gemini) and US\$6.80 (GPT-image) with one runner per output directory.
P1s reads: ``This photo shows three flat panels standing in a row, about \{distances\}. Edit it so it looks like it was taken with a full-frame camera (36 mm wide sensor) and a 50mm lens at f/$N$ instead of the all-in-focus original, from the same camera position, focused on the \{focus panel\}. Keep the composition and the content of every panel exactly as it is.''
Both editors returned all 120 requested images without a failed or duplicate call (adapter-estimated cost US\$8.04 and US\$6.49). Panels with coverage below 20\% are excluded from these indices (prospective rule); the sensitivity analysis including them gives the same classifications.

Usable panels (accepted, at least 20\% coverage) in layouts R and F, out of 24 per cell, by panel and aperture (f/1.4 | f/2.8): Gemini R left 21 | 20, centre 16 | 14, right 23 | 24; Gemini F left 24 | 24, centre 19 | 23, right 23 | 20; GPT-image R left 22 | 22, centre 14 | 12, right 24 | 24; GPT-image F left 24 | 24, centre 6 | 10, right 20 | 21.
The depth and aperture indices need both blurred panels: 18 (Gemini R), 16 (Gemini F), 20 (GPT-image R), and 1 (GPT-image F) of 24 jobs have both blurred panels usable at both apertures. The registered statistic pools medians over available panels, so these counts bound how much any single job contributes; GPT-image's layout-F depth index rests mainly on 16 centre measurements.

\subsection{Intervention blocks}
Each block below was registered (frozen text with timestamp and hash) before its first editor call, used six scenes unless stated, one retained output per cell, one runner per output directory, and logged every attempt; none had a failed or duplicate call. Adapter-estimated cost (US\$): v4 5.81 (48 cells per editor, extreme distances); v5 4.36 (36); v6 5.81 (48); v7 2.91 (24); v8 4.36 (36); v9a 1.45 (12); v9b 1.45 (12); v9c 2.90 (24); v10 4.35 (36); v11 2.90 (24).
Decision gates: v8 and v9 were run only because v7 did not support the straddle observation; the order of the night run was fixed in advance.
Deviations: in v9b the analysis script contained an extra clause labelling an interval that excludes 0 as dependent (not triggered); in v11 the widths were floored at 0.4 px before taking ratios because a deep-depth-of-field median can be exactly 0; the night driver's own cost total omitted the v10 ledger, which the totals above include.
Prompts: v4--v8 use P1s with the stated distances of the geometry; v9a asks for focus on the right panel at f/1.4; v9b removes floor and wall from the render; v9c asks for ``a Gaussian blur with a standard deviation of about 2 (6) pixels'' with the centre panel sharp; v10 uses the wordings ``shot wide open with a shallow depth of field at f/$N$ \dots with natural lens blur and bokeh'', a Chinese sentence naming the 50\,mm lens, f/$N$ and the focus on the middle panel, and ``Change the aperture to f/$N$.''; v11 asks for a ``very shallow depth of field (strong background blur)'' or a ``very deep depth of field (everything nearly sharp)'' with distances and focus stated and no f-number.

\subsection{Geometries of the intervention blocks}
\Cref{tab:geoms} lists the panel distances and the thin-lens left/right width ratio fixed in the registration of each geometry; all other blocks (v5, v6, v9, v10, v11) use the original geometry of 1.5, 2.2, and 3.0\,m, with the focus on the centre panel unless the block says otherwise (v6: focus on the left panel at 1.5\,m, or the right panel at 3.0\,m; v9a: focus on the right panel).
Geometries G1, G2, S1, S2, and those of v8 use panels of equal projected size (340\,px) and a focus distance of 2.2\,m; the measured oracle ratios differ from the design values for G1 (0.10) and G2 (5.8) and are 2.59, 2.62, and 2.61 for the three v8 conditions.

\begin{table}[t]
\centering\scriptsize\setlength{\tabcolsep}{2pt}
\caption{Geometries of the intervention blocks (left, centre, right panel distances in m; design left/right thin-lens width ratio at registration; the oracle's measured ratio is quoted in the text above where it differs).}
\label{tab:geoms}
\begin{tabular}{llcc}
\toprule
Block & Geometry & L / C / R (m) & design ratio \\
\midrule
v4 & G1 (far much farther) & 2.0 / 2.2 / 6.0 & 0.16 \\
v4 & G2 (near much closer) & 1.4 / 2.2 / 2.6 & 3.7 \\
v7 & S1 (both behind focus) & 3.0 / 2.2 / 6.0 & 0.42 \\
v7 & S2 (both in front) & 1.2 / 2.2 / 1.8 & 3.75 \\
v8 & O (opposite sides) & 1.53 / 2.2 / 2.67 & 2.5 \\
v8 & SF (both in front) & 1.53 / 2.2 / 1.87 & 2.5 \\
v8 & SB (both behind) & 3.93 / 2.2 / 2.67 & 2.5 \\
\bottomrule
\end{tabular}
\end{table}

\subsection{Exploratory exponent summary}
This summary is post hoc and was not registered.
Gemini's near/far ratio equals the oracle's ratio to a power of 0.21--0.60 in the nine conditions with the focus on the centre panel (original, reversed, S1, S2, G1, G2, and the three conditions of v8), and to 0.74 and 1.07 in the two conditions with the focus on a side panel; GPT-image's exponent is smaller and differs between configurations.
It condenses the compression into one number per condition.

\subsection{Calibration grid}
We imposed known blur on the left panel of 12 jobs and measured it with the same estimator, at native size and at Gemini's size with JPEG q90. Recovered over imposed width, Gaussian kernel: 0.68 (native) and 0.93 (Gemini size) at 0.5 px, 1.07 and 1.19 at 1 px, and 1.00--1.01 (native) and 1.00--1.06 (Gemini size) from 1.5 to 8 px.
Disc kernel with Gaussian-equivalent standard deviation $s$: 0.86 (native) and 1.69 (Gemini size) at 0.5 px, 1.17 and 1.26 at 1 px, 1.11--1.25 from 1.5 to 6 px at both sizes, and saturation at the estimator range for 8 px (10 of 12 jobs measurable).
For the tested native left-panel medians at imposed widths 2 and 6 px, a disc kernel is recovered as 2.22 and 7.52 px, a ratio of 3.39 against 3.00 (12.9\% inflation); the grid tests the left panel only and does not establish a general two-panel bound.

\subsection{Analytic and slanted-edge cross-checks}
\paragraph{Analytic.} For every oracle render of the original, extreme-distance, same-side, matched-width and moved-focus geometries we computed the thin-lens circle of confusion in pixels, $c=(f^{2}/N)\,z_f/(z_f-f)\,|1/z-1/z_f|\,(1536/36)$, and its disc standard deviation $c/4$, and compared it with the estimator's median reading over jobs.
For the 333 of 432 points with $c/4\ge 1$\,px, the ratio measured/analytic has median 1.17 (interquartile range 1.14--1.19, range 1.01--1.32): 1.15 for 1--2 px, 1.17 for 2--4 px, 1.22 for 4--8 px; below 1 px it is 0.77 (99 points).
\paragraph{Slanted edge.} We rendered a target with a high-contrast edge tilted 4--6$^\circ$ from vertical and speckle texture on all three panels (true thin-lens depth of field, f/1.4, 2.8, 5.6) and measured the standard deviation of the line-spread function from the edge-spread function (0.25 px bins; slanted-edge procedure after \citealp{williams2014slanted}), after removing each panel's sharp-render floor in quadrature.
On known Gaussian blur imposed on the target the method returns 0.93--1.06 of the imposed width from 1 to 6 px (native size; Gemini size 0.94--1.13), and on synthetic disc blur 0.87--0.97 of the disc's standard deviation, while the estimator returns 1.00--1.02 (Gaussian, $\ge$2 px) and 1.12--1.28 (disc) of the imposed width.
On the rendered thin-lens blur of the near panel the edge method gives 3.19 px (f/1.4) and 1.39 px (f/2.8), against 4.13 and 2.07 px analytically (0.77 and 0.67), and the estimator reads 4.99 and 2.36 px; the estimator-to-edge ratio is therefore 1.57 and 1.70 on rendered blur against 1.24--1.35 on synthetic discs.
The far panel's sharp-render floor (2.7 px with this texture) exceeds its blur, so the edge method is unusable there.
Reading: the analytic agreement (1.17) partly reflects two offsetting differences (the rendered blur is narrower than the formula and the estimator reads rendered blur higher than a synthetic disc of the same width), so the absolute scale of the Gaussian-equivalent width is uncertain by about 25\%; ratios of readings of the same estimator, which all indices use, cancel a common scale factor, while any width-, texture-, or kernel-dependent bias is bounded only by the calibration grid and the controls. Not resolved: whether the renderer's kernel differs from a uniform disc, and the editors' unknown kernels.

\subsection{Data and licensing}
Panel textures in the stimuli and figures are crops of RealBokeh f/22 images \citep{seizinger2025bokehlicious}, released under CC BY-NC-SA 4.0; figures and stimuli that contain them are shared under the same license and for non-commercial research use.
The centre-panel mesh and the floor and wall textures are generated by us.

\subsection{Preservation floors}
Median panel widths of the 12 ``return unchanged'' images (left/centre/right, px): Gemini 0.38/0.94/0.53 and GPT-image 0.36/0.98/0.34.
The mesh centre panel has a floor near 1\,px for both editors.
Resizing an unchanged all-in-focus render to Gemini's output size alone gives a centre width near 0.37\,px.

\subsection{Estimator details}
Working resolution: 1000\,px image width. Gaussian bank: 0 and 0.4--6.0\,px in steps of 0.4\,px (at working resolution). Windows of 41\,px are centred at every interior pixel, high-passed by subtracting a Gaussian of width 5\,px, restricted to texture above the 40th percentile of the reference, and measurable if the peak correlation is at least 0.80; a panel width is reported only when at least 30 positions are measurable.
Panel crops exclude 16\,px at the 1536\,px image width, followed by an additional 8\,px margin at the 1000\,px working width.
Oracle width medians (px): f/1.4 near 5.01 and far 2.76; f/2.8 near 2.35 and far 1.40; f/5.6 near 1.16 and far 0.43.
Three far-panel oracle measurements at each of f/1.4 and f/2.8 are rejected by the alignment rule, so the oracle denominators use 45 of 48 far panels at these apertures.
Coverage is the fraction of interior window positions with a measurable width. For accepted, measurable far panels at the primary apertures:
\begin{center}\footnotesize\setlength{\tabcolsep}{3pt}\begin{tabular}{@{}lccc@{}}\toprule Stratum & min cov. & median cov. & below 20\% \\\midrule Gemini P1 & 0.57\% & 60.27\% & 3/85 \\ Gemini P2 & 5.76\% & 60.32\% & 1/82 \\ GPT-image P1 & 1.51\% & 60.34\% & 4/90 \\ GPT-image P2 & 2.56\% & 60.50\% & 3/84 \\\bottomrule\end{tabular}\end{center}
The 20\% summary is descriptive; no coverage exclusion is applied to the registered endpoints. The lowest-coverage panel has 138 of 24066 positions (0.57\%).

\subsection{Missing-panel sensitivity}
Every rejected or unmeasurable panel is included. The estimator's range is 0--9.216\,px at the 1536\,px image width. To maximize the depth index, missing near widths are assigned 9.216\,px and missing far widths 0\,px; to maximize the aperture index, missing f/1.4 widths are assigned 9.216\,px and missing f/2.8 widths 0\,px. Pooled medians and the 4000-draw scene bootstrap are then recomputed with the measured oracle fixed. Resulting upper bounds: depth 0.529 and 0.540 (Gemini P1, P2) and 0.215 and 0.130 (GPT-image P1, P2); aperture 0.255, 0.333, 0.283, 0.291 (Gemini P1, P2, GPT-image P1, P2).

%% file: references.bib
@misc{rueckerl2026camera,
  doi = {10.48550/ARXIV.2609.37732},
  url = {https://arxiv.org/abs/2609.37732},
  author = {Rückerl, Sebastian},
  title = {The Camera Inside the Editor: Reading the Implicit Camera of Image Editors with Painted Calibration Patterns},
  publisher = {arXiv},
  year = {2026},
  copyright = {Creative Commons Attribution 4.0 International}
}

@inproceedings{shen2025defocus,
  author = {Shrivastava, Ayush and Barnes, Connelly and Zhang, Xuaner and Zhang, Lingzhi and Owens, Andrew and Amirghodsi, Sohrab and Shechtman, Eli},
  title = {Fine-grained Defocus Blur Control for Generative Image Models},
  booktitle = {2026 IEEE/CVF Winter Conference on Applications of Computer Vision (WACV)},
  year = {2026},
  doi = {10.1109/WACV61042.2026.00422}
}

@inproceedings{fortes2025bokeh,
  author = {Fortes, Armando and Wei, Tianyi and Zhou, Shangchen and Pan, Xingang},
  title = {Bokeh Diffusion: Defocus Blur Control in Text-to-Image Diffusion Models},
  booktitle = {Proceedings of the SIGGRAPH Asia 2025 Conference Papers},
  year = {2025},
  doi = {10.1145/3757377.3763906}
}

@misc{mu2025refocus,
  doi = {10.48550/ARXIV.2512.16923},
  url = {https://arxiv.org/abs/2512.16923},
  author = {Mu, Chun-Wei Tuan and Fan, Cheng-De and Huang, Jia-Bin and Liu, Yu-Lun},
  title = {Generative Refocusing: Flexible Defocus Control from a Single Image},
  publisher = {arXiv},
  year = {2025},
  copyright = {Creative Commons Attribution 4.0 International}
}

@misc{anybokeh2026,
  doi = {10.48550/ARXIV.2606.31959},
  url = {https://arxiv.org/abs/2606.31959},
  author = {Hou, Xinyu and Li, Xiaoming and Yue, Zongsheng and Loy, Chen Change},
  title = {AnyBokeh: Physics-Guided Any-to-Any Bokeh Editing with Optical Fingerprint Transfer},
  publisher = {arXiv},
  year = {2026},
  copyright = {Creative Commons Attribution 4.0 International}
}

@misc{lensstyle2026,
  doi = {10.48550/ARXIV.2609.04939},
  url = {https://arxiv.org/abs/2609.04939},
  author = {Huang, Yachuan and Xiao, Liwen and Shen, Liao and Wang, Qiwen and Sun, Huiqiang and Pan, Zhiyu and Cao, Zhiguo},
  title = {LensStyle: Learning the Optical Aesthetics for Controllable Stylized Lens Effect Rendering},
  publisher = {arXiv},
  year = {2026},
  copyright = {Creative Commons Attribution 4.0 International}
}

@misc{kuechler2026lighting,
  doi = {10.48550/ARXIV.2606.26738},
  url = {https://arxiv.org/abs/2606.26738},
  author = {Küchler, Tim and Feiden, Johann-Friedrich and Nießner, Matthias and Rother, Carsten},
  title = {Do Image Editing Models Understand Lighting?},
  publisher = {arXiv},
  year = {2026},
  copyright = {Creative Commons Attribution 4.0 International}
}

@misc{picabench2025,
  doi = {10.48550/ARXIV.2510.17681},
  url = {https://arxiv.org/abs/2510.17681},
  author = {Pu, Yuandong and Zhuo, Le and Han, Songhao and Xing, Jinbo and Zhu, Kaiwen and Cao, Shuo and Fu, Bin and Liu, Si and Li, Hongsheng and Qiao, Yu and Zhang, Wenlong and Chen, Xi and Liu, Yihao},
  title = {PICABench: How Far Are We from Physically Realistic Image Editing?},
  publisher = {arXiv},
  year = {2025},
  copyright = {arXiv.org perpetual, non-exclusive license}
}

@misc{phyeditbench2026,
  doi = {10.48550/ARXIV.2606.26551},
  url = {https://arxiv.org/abs/2606.26551},
  author = {Guo, Shengbin and He, Shaokang and Meng, Chaoyue and Xiao, Shengpeng and Xiang, Xunzhi and Zhang, Shaofeng and Fan, Qi},
  title = {PhyEditBench: A Real-World Multi-Stage Benchmark for Physics-Aware Image Editing},
  publisher = {arXiv},
  year = {2026},
  copyright = {arXiv.org perpetual, non-exclusive license}
}

@misc{spatialedit2026,
  doi = {10.48550/ARXIV.2604.04911},
  url = {https://arxiv.org/abs/2604.04911},
  author = {Xiao, Yicheng and Zhang, Wenhu and Song, Lin and Chen, Yukang and Li, Wenbo and Jiang, Nan and Ren, Tianhe and Lin, Haokun and Huang, Wei and Huang, Haoyang and Li, Xiu and Duan, Nan and Qi, Xiaojuan},
  title = {SpatialEdit: Benchmarking Fine-Grained Image Spatial Editing},
  publisher = {arXiv},
  year = {2026},
  copyright = {arXiv.org perpetual, non-exclusive license}
}

@misc{seizinger2025bokehlicious,
  doi = {10.48550/ARXIV.2503.16067},
  url = {https://arxiv.org/abs/2503.16067},
  author = {Seizinger, Tim and Vasluianu, Florin-Alexandru and Conde, Marcos V. and Wu, Zongwei and Timofte, Radu},
  title = {Bokehlicious: Photorealistic Bokeh Rendering with Controllable Apertures},
  publisher = {arXiv},
  year = {2025},
  copyright = {Creative Commons Attribution Non Commercial Share Alike 4.0 International}
}

@inproceedings{drbokeh2024,
  author = {Sheng, Yichen and Yu, Zixun and Ling, Lu and Cao, Zhiwen and Zhang, Xuaner and Lu, Xin and Xian, Ke and Lin, Haiting and Benes, Bedrich},
  title = {Dr.Bokeh: DiffeRentiable Occlusion-aware Bokeh Rendering},
  booktitle = {Proceedings of the IEEE/CVF Conference on Computer Vision and Pattern Recognition (CVPR)},
  year = {2024},
  url = {https://openaccess.thecvf.com/content/CVPR2024/papers/Sheng_Dr._Bokeh_DiffeRentiable_Occlusion-aware_Bokeh_Rendering_CVPR_2024_paper.pdf}
}

@misc{hpedit2026,
  doi = {10.48550/ARXIV.2604.19406},
  url = {https://arxiv.org/abs/2604.19406},
  author = {Li, Fan and Wang, Chonghuinan and Lei, Lina and Qiu, Yuping and Xu, Jiaqi and Jiang, Jiaxiu and Qin, Xinran and Chen, Zhikai and Song, Fenglong and Wang, Zhixin and Pei, Renjing and Zuo, Wangmeng},
  title = {HP-Edit: A Human-Preference Post-Training Framework for Image Editing},
  publisher = {arXiv},
  year = {2026},
  copyright = {Creative Commons Attribution 4.0 International}
}

@inproceedings{qin2025camedit, series={NeurIPS 2025}, title={CamEdit: Continuous Camera Parameter Control for Photorealistic Image Editing}, url={http://dx.doi.org/10.52202/085713-3808}, DOI={10.52202/085713-3808}, booktitle={Advances in Neural Information Processing Systems 38}, publisher={Neural Information Processing Systems Foundation, Inc. (NeurIPS)}, author={Qin, Xinran and Wang, Zhixin and Li, Fan and Chen, Haoyu and Pei, Renjing and Li, Wenbo and Cao, Xiaochun}, year={2025}, pages={126619–126638}, collection={NeurIPS 2025} }

@article{evangelidis2008ecc, title={Parametric Image Alignment Using Enhanced Correlation Coefficient Maximization}, volume={30}, ISSN={0162-8828}, url={http://dx.doi.org/10.1109/TPAMI.2008.113}, DOI={10.1109/tpami.2008.113}, number={10}, journal={IEEE Transactions on Pattern Analysis and Machine Intelligence}, publisher={Institute of Electrical and Electronics Engineers (IEEE)}, author={Evangelidis, G.D. and Psarakis, E.Z.}, year={2008}, month=Oct, pages={1858–1865} }

@article{subbarao1994dfd, title={Depth from defocus: A spatial domain approach}, volume={13}, ISSN={1573-1405}, url={http://dx.doi.org/10.1007/bf02028349}, DOI={10.1007/bf02028349}, number={3}, journal={International Journal of Computer Vision}, publisher={Springer Science and Business Media LLC}, author={Subbarao, Murali and Surya, Gopal}, year={1994}, month=Dec, pages={271–294} }

@article{zhuo2011defocus, title={Defocus map estimation from a single image}, volume={44}, ISSN={0031-3203}, url={http://dx.doi.org/10.1016/j.patcog.2011.03.009}, DOI={10.1016/j.patcog.2011.03.009}, number={9}, journal={Pattern Recognition}, publisher={Elsevier BV}, author={Zhuo, Shaojie and Sim, Terence}, year={2011}, month=Sept, pages={1852–1858} }

@inproceedings{hauser2018bokeh,
  author = {Hauser, Wolf and Neveu, Benoit and Jourdain, Jean-Benoit and Viard, Cl{\'e}ment and Guichard, Fr{\'e}d{\'e}ric},
  title = {Image quality benchmark of computational bokeh},
  booktitle = {Electronic Imaging},
  year = {2018},
  doi = {10.2352/ISSN.2470-1173.2018.12.IQSP-340}
}

@misc{zhang2025bokehdepth,
  author = {Zhang, Hangwei and Fortes, Armando and Wei, Tianyi and Pan, Xingang},
  title = {Boosting Monocular Metric Depth Estimation via Bokeh Rendering},
  year = {2025},
  howpublished = {arXiv:2512.12425},
  doi = {10.48550/arXiv.2512.12425}
}

@inproceedings{potmesil1981lens,
  author = {Potmesil, Michael and Chakravarty, Indranil},
  title = {A lens and aperture camera model for synthetic image generation},
  booktitle = {Proceedings of the 8th Annual Conference on Computer Graphics and Interactive Techniques (SIGGRAPH)},
  year = {1981},
  pages = {297--305},
  doi = {10.1145/800224.806818}
}

@inproceedings{williams2014slanted,
  author = {Williams, Don and Burns, Peter D.},
  title = {Evolution of slanted edge gradient {SFR} measurement},
  booktitle = {Image Quality and System Performance XI (SPIE Proceedings)},
  volume = {9016},
  year = {2014},
  doi = {10.1117/12.2042706}
}
